\documentclass[11pt,a4paper]{article}
\usepackage{times,latexsym}
\usepackage{url}
\usepackage[T1]{fontenc}

\def\doubleunderline#1{\underline{\underline{#1}}}

\usepackage{microtype}

\usepackage{dirtytalk} %
\usepackage{booktabs}
\usepackage{amsfonts}
\usepackage{amsmath}
\usepackage{enumitem}
\usepackage{makecell}
\usepackage{diagbox}
\usepackage{multirow}

\usepackage{subcaption}

\usepackage{algorithm}%
\usepackage[noend]{algpseudocode}%

\usepackage[pdftex]{graphicx}

\DeclareMathOperator*{\dec}{dec}

\usepackage[acceptedWithA]{tacl2021v1}
\usepackage{xspace,mfirstuc,tabulary}

\newif\iftaclinstructions
\taclinstructionsfalse %
\iftaclinstructions
\renewcommand{\confidential}{}
\renewcommand{\anonsubtext}{(No author info supplied here, for consistency with
TACL-submission anonymization requirements)}
\newcommand{\instr}
\fi

\iftaclpubformat %

\else

\fi

\title{An Analysis of BPE Vocabulary Trimming \\ in Neural Machine Translation}

\author{Marco Cognetta \\
  Department of Computer Science \\
  Tokyo Institute of Technology \\
  \texttt{cognetta.marco@gmail.com} \\\And
  Tatsuya Hiraoka \\
  Fujitsu Limited \\
  \texttt{hiraoka.tatsuya@fujitsu.com} \\\ \AND
  Naoaki Okazaki \\
  Department of Computer Science \\
  Tokyo Institute of Technology \\
  \texttt{okazaki@c.titech.ac.jp} \\\And
  Rico Sennrich \\
  Department of Computational Linguistics \\
  University of Zurich  \\
  \texttt{sennrich@cl.uzh.ch} \\\ \AND
  Yuval Pinter \\
  Department of Computer Science \\
  Ben-Gurion University of the Negev \\
  \texttt{uvp@cs.bgu.ac.il} \\}

\date{}

\begin{document}
\maketitle
\begin{abstract}

We explore threshold vocabulary trimming in Byte-Pair Encoding subword tokenization, a postprocessing step that replaces rare subwords with their component subwords.
The technique is available in popular tokenization libraries but has not been subjected to rigorous scientific scrutiny.
While the removal of rare subwords is suggested as best practice in machine translation implementations, both as a means to reduce model size and for improving model performance through robustness, our experiments indicate that, across a large space of hyperparameter settings, vocabulary trimming fails to improve performance, and is even prone to incurring heavy degradation.

\end{abstract}

\section{Introduction}

\begin{figure*}[t]
    \centering
    \includegraphics[width=1.5\columnwidth]{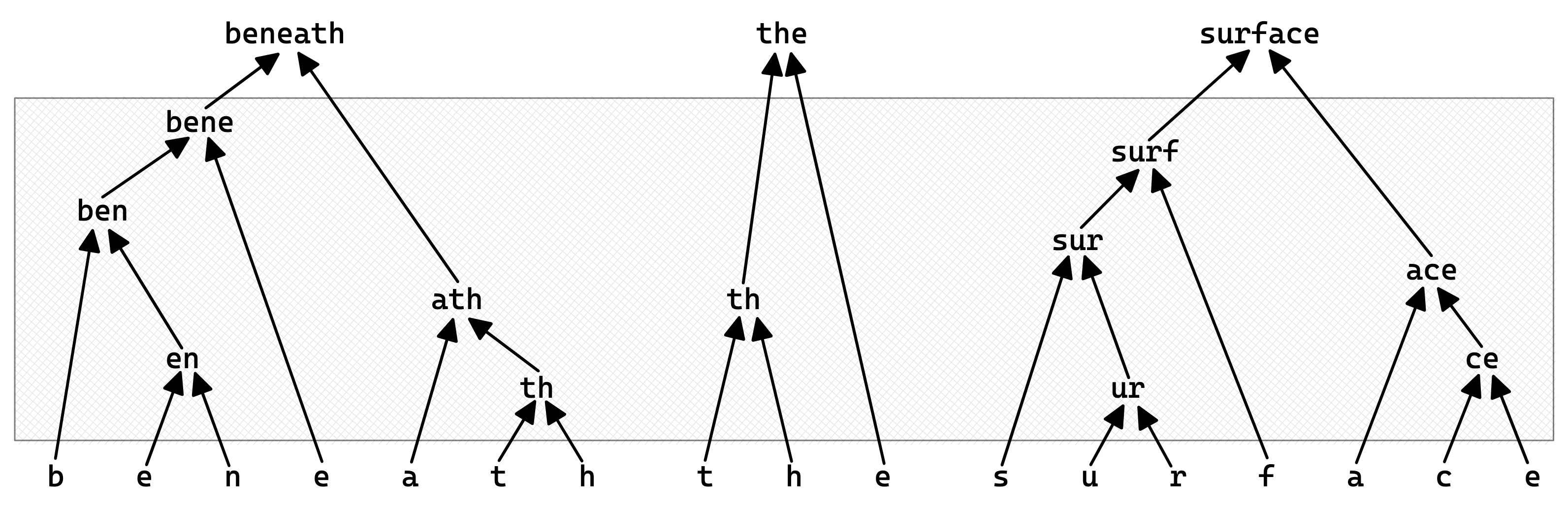}
    \caption{An example BPE tokenization. The shaded region contains \textit{intermediate} subwords, which often appear during the merging process to build longer subwords, but rarely in the final tokenization of a word.
    \label{fig:sentence_example}}
\end{figure*}

Subword tokenization is an important process in modern neural language modeling, as it enables models to represent any possible word over a known alphabet while keeping the vocabulary size small.
One of the most common subword tokenization methods %
is Byte-Pair Encoding~\cite[BPE;][]{Gage1994ANA,sennrich-etal-2016-neural}, a greedy, statistical subword tokenization method popular particularly in machine translation applications.
BPE builds its vocabulary and tokenizes a corpus by iteratively replacing the most frequently co-occurring token pair with a single merged token.
An unfortunate side-effect of this process is the existence of \say{intermediate} subwords---subwords that only appear during the process of forming longer subwords, and rarely appear as output tokens in the final sequence, as shown in \autoref{fig:sentence_example}.

Vocabulary trimming is a tokenization post-processing step where subwords that appear fewer than a prescribed number of times in a given corpus are replaced with their component subwords.
This technique is recommended as a best practice when implementing BPE-based machine translation models, and stems from the widespread practice of learning a joint BPE vocabulary for both the source and target languages~\cite{subword-nmt, sennrich-etal-2016-neural, sennrich-etal-2017-university}.
In such settings, many subwords that appear only in the source or only in the target language would be present in the vocabulary for both.
However, this technique can also be used to eliminate rare or intermediate subwords, which do not appear enough during training for the model to learn robust representations of them, and may degrade the downstream performance of the model~\cite{sennrich-etal-2017-university, sennrich-zhang-2019-revisiting}.
In addition, the removal of tokens leads to immediate savings in parameter budgets in the token embedding layer.
While the intuition behind trimming is straightforward, it has never been evaluated via controlled experimentation.

We present a comprehensive study aimed at understanding the actual effect of vocabulary trimming on the performance of machine translation systems.
Among the settings we evaluate are\footnote{Code to reproduce all experiments will be made available.}:
\begin{enumerate}[label={\arabic*)}]
    \item Trimming an optimal baseline (Section~\ref{ssec:optimalbaselinetrimming})
    \item Whether trimming helps recover some performance in a suboptimal baseline (Section~\ref{ssec:suboptimalbaselinetrimming})
    \item Trimming only the source vocabulary vs. only that of the target language (Section~\ref{ssec:sourcevstarget})
    \item Trimming by a percentile-frequency heuristic~\cite{gowda-may-2020-finding} (Section~\ref{ssec:100heuristic})
    \item The effect of trimming on performance over sentences with rare subwords (Section~\ref{ssec:raresentences})
    \item Trimming, but preserving subwords that do not appear in a larger merge (Section~\ref{ssec:terminalpreservation})
    \item Trimming compared to using a smaller vocabulary (Section~\ref{ssec:stunted})
    \item Using a joint vocabulary (Section~\ref{ssec:joint_vocab})
    \item Repeating (2) but with a larger dataset (Section~\ref{ssec:larger_dataset})
    \item Initializing an extremely large base vocabulary and trimming (Section~\ref{ssec:large})
\end{enumerate}

In general, for our setting of BPE tokenization, we find that vocabulary trimming has no positive effect on model quality, and in many cases can substantially degrade it.

\section{Byte-Pair Encoding}

BPE is a commonly-used subword vocabulary generation and tokenization method for neural language modeling.
A BPE tokenizer is built from corpus statistics using a greedy merging scheme, as described in Algorithm \ref{alg:bpe}.
A subword vocabulary $\mathcal{V}$ is built by iteratively merging the token pairs with highest frequency in the corpus, starting from all individual character sequences.
When a token pair $p = (l, r)$ is merged, the merge information $(p, (l, r))$ is added to an ordered list of merges $\mathcal{M}$, $p$ is added to $\mathcal{V}$, and every instance of $l\circ r$ in the corpus is replaced with $p$.
Crucially, tokens are never removed from the vocabulary.

In the standard application of BPE tokenization, given a trained vocabulary in the form of a merge list, each word is considered individually.
Starting from the character sequence of the word, the highest ranked token pair in $\mathcal{M}$ is merged.
The same rule is then applied recursively to the output sequence, and so on until there are no more valid merges remaining.
In the case of frequent words, this typically culminates with the entire word becoming a single token.

\begin{algorithm}[t]
    \small
    \begin{algorithmic}[1]
        \Procedure{BPEInit}{Corpus $\mathcal{C}$, Vocab Size $n$}
            \State Initialize $\mathcal{V}$ with all characters in $\mathcal{C}$
            \State Initialize $\mathcal{M}$ as an empty list
            \While{$|\mathcal{V}| \le n$}
                \State $p \gets \texttt{argmax}_{(l,r) \in \mathcal{V}\times\mathcal{V}}~\texttt{count}(l\circ r, \mathcal{C})$
                \State $\mathcal{V} \gets \mathcal{V} \cup \{p\}$
                \State $\mathcal{C} \gets \texttt{mergeall}(\mathcal{C}, (l\circ r), p)$
                \State $\mathcal{M} \gets \texttt{append}(\mathcal{M}, (p, (l, r)))$
            \EndWhile
            \State \textbf{return} $\mathcal{V}, \mathcal{M}$
        \EndProcedure
    \end{algorithmic}
\caption{BPE Vocabulary Construction}\label{alg:bpe}
\end{algorithm}

\begin{figure*}[ht]
    \centering
    \includegraphics[width=1.6\columnwidth]{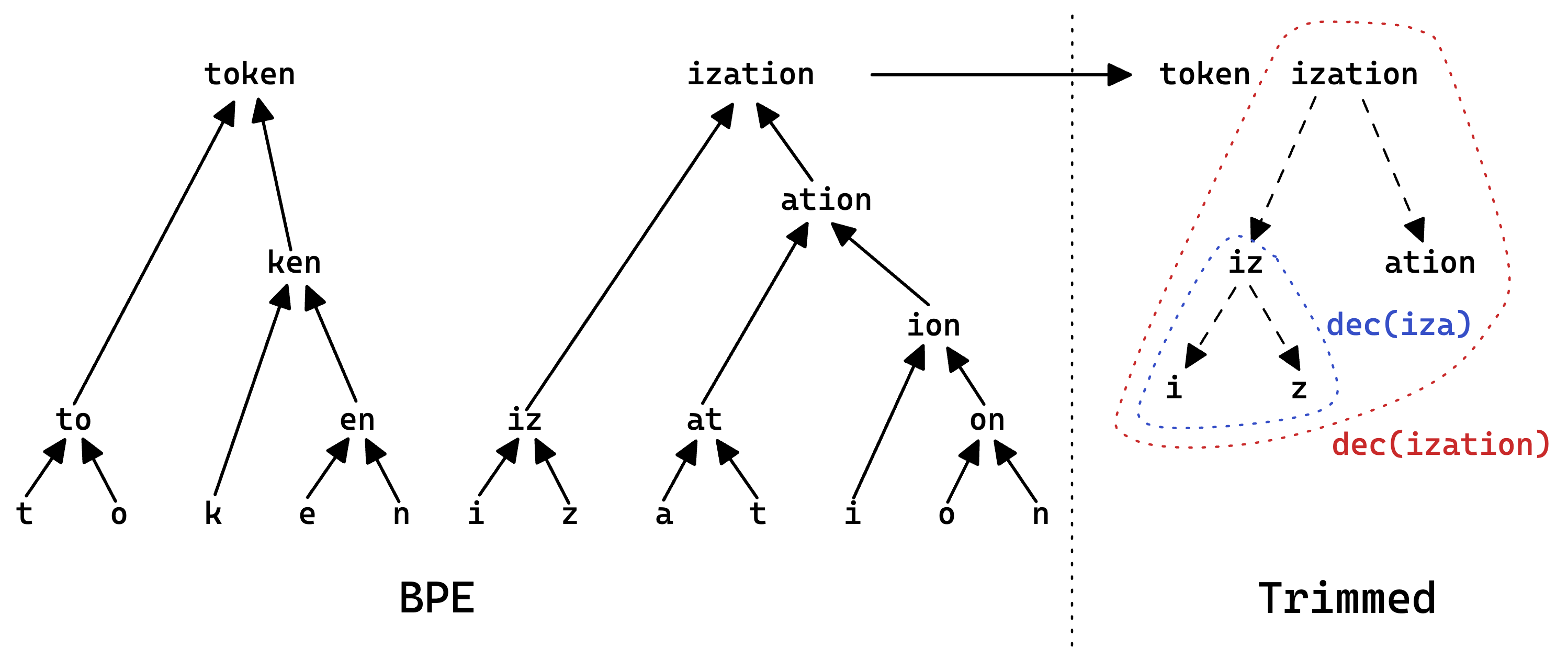}
    \caption{An example of a trimmed tokenizer during inference. The left side shows the original character sequence and the final BPE-tokenized sequence, \texttt{token·ization}. On the right, tokens are decomposed if they have less than the designated threshold, according to the function \textsc{dec}. The tokens \texttt{ization} = (\texttt{iz}, \texttt{ation}) and \texttt{iz} = (\texttt{i}, \texttt{z}) are in $\mathcal{X}_{\mathcal{B}, T}$ and are decomposed, resulting in \texttt{token·i·z·ation}.\label{fig:trimmedbpedecomposition}}
\end{figure*}

While the BPE vocabulary construction algorithm focuses on frequency-based optimization within each step in the main loop, this \emph{local} property causes a major failure on the \emph{global} level, creating vocabularies that contain many infrequent tokens.
This can cause many parameters of the model to be occupied by unused or poorly-trained tokens, potentially reducing its performance.
The root cause for this behavior is that in natural language, many frequent tokens are long.
In order for BPE to form long subwords, it must first add many shorter subwords to the vocabulary so that they can be merged into a larger one.
These smaller subwords, by definition, appear frequently at the time that they are introduced into the vocabulary, but once they are used to form larger tokens, they may never appear again outside a further mergeable environment.
For example, consider the (sub)word \texttt{Kentucky} being formed by merging the subword pair \texttt{Kentuc·ky}.
The subword pair \texttt{Kentuc·ky} was, at the time that it was added to the vocabulary, the most frequently co-occuring subword pair in the corpus.
However, after \texttt{Kentucky} is formed and added to the vocabulary, the subword \texttt{Kentuc}, which does not appear in any other words in the corpus and always appears directly before a \texttt{ky} subword, will never appear in the final tokenization of a word and will only ever be used to eventually form the subword \texttt{Kentucky}.
It is rarely-occuring or intermediate subword tokens like \texttt{Kentuc}, which only appear on the path to forming longer subwords, that we seek to remove from the downstream model's vocabulary.

\subsection{Joint Vocabulary Construction}
In neural machine translation, the BPE vocabulary is often learned jointly over both source and target languages.
In practice, this is done by simply concatenating the corpora, allowing languages that share common words to have one-to-one alignment of the tokens~\cite{sennrich-etal-2016-neural}.
In many cases, the pessimal case being where the source and target languages do not even share a common alphabet, there can be subwords that appear in only one language or the other, but are present in both of their vocabularies due to the joint training.
It was for this reason that vocabulary trimming was originally introduced, as one could easily remove all subwords that appeared only in one corpus to reduce the final model size without sacrificing performance~\cite{sennrich-etal-2017-university, sennrich-zhang-2019-revisiting}.
In this paper, we only consider the split-vocabulary setting, except in Section \ref{ssec:joint_vocab}, which focuses on the joint-vocabulary setting.

\section{Vocabulary Trimming}

Vocabulary trimming is a simple procedure built on top of a BPE tokenizer.
Let $\mathcal{B} = (\mathcal{V_B}, \mathcal{M_B})$ be a BPE tokenizer trained on corpus $\mathcal{C}$.
$\mathcal{B}$ defines a function, $\mathcal{B}: \Sigma_\mathcal{B}^+ \rightarrow \mathcal{V_B^+}$, that maps character sequences into subword token sequences, where $\Sigma_{\mathcal{B}} \subseteq \mathcal{V_B}$ is the set of \textit{atomic} characters.
For every $v \in \mathcal{V}_{\mathcal{B}}\setminus\Sigma_{\mathcal{B}}$, let $(l_v, r_v)$ be the subwords that formed $v$ during the creation process,
and let $c_v$ be the number of times a token $v$ appears in the tokenized version of $\mathcal{C}$.
Given a threshold $\mathbb{T} \in \mathbb{N}$, let
$\mathcal{X}_{\mathcal{B}, \mathbb{T}} = \{ v \in \mathcal{V_B}\setminus\Sigma_{\mathcal{B}} \mid c_v \le \mathbb{T} \}$
be the set of non-atomic subword tokens that appear at most $\mathbb{T}$ times in the corpus after being tokenized by $\mathcal{B}$.

Next, let  $\dec_{\mathcal{X}_{\mathcal{B}, \mathbb{T}}}: \mathcal{V_B} \rightarrow \mathcal{V_B^+}$ be a recursive decomposition function:
{\small
\[
  \dec_{\mathcal{X}_{\mathcal{B}, \mathbb{T}}}(v) =  \begin{cases}
        v & \text{if } v \notin \mathcal{X}_{\mathcal{B}, \mathbb{T}}\\
        \dec_{\mathcal{X}_{\mathcal{B}, \mathbb{T}}}(l_v)\circ \dec_{\mathcal{X}_{\mathcal{B}, \mathbb{T}}}(r_v) & \text{otherwise.}
    \end{cases}
\]}

In words, $\dec_{\mathcal{X}_{\mathcal{B}, \mathbb{T}}}$ recursively decomposes a subword $v$ into its component subwords, until the remaining subwords all appear more than $\mathbb{T}$ times in the corpus or are atomic characters.

Given a BPE tokenizer $\mathcal{B}$, a trimmed tokenizer $\mathcal{B}'$ has a final subword vocabulary $\mathcal{V_{B'}} = \mathcal{V_B} \setminus \mathcal{X}_{\mathcal{B}, \mathbb{T}}$.
Note that $\Sigma_{\mathcal{B}} = \Sigma_{\mathcal{B'}}$ and $\mathcal{M}_\mathcal{B} = \mathcal{M}_\mathcal{B'}$.
In order to tokenize an input sequence $z \in \Sigma_{\mathcal{B'}}^+$, a trimmed tokenizer computes:
\[\mathcal{B'}(z) = \dec_{\mathcal{X}_{\mathcal{B}, \mathbb{T}}}(\mathcal{B}(z)),\] 
which is the decomposition of the $\mathcal{B}$-tokenized sequence, according to $\dec_{\mathcal{X}_{\mathcal{B}, \mathbb{T}}}$.
\autoref{fig:trimmedbpedecomposition} provides an example of a word being tokenized by $\mathcal{B}$ and then decomposed, where appropriate, by $\mathcal{B'}$.

\section{Experiments}

To determine the effect of vocabulary trimming, we use the IWSLT14 German\textrightarrow English translation task~\cite{cettolo-etal-2014-report} and experiment with varying the source and target BPE vocabulary sizes\footnote{Due to how \texttt{subword-nmt} produces the vocabulary, the final effective vocabulary size is not always exactly equal to the desired size, but the difference is typically very small.}, %
$\mathbb{B}_{s}$ and $\mathbb{B}_t$, and the source and target thresholds, $\mathbb{T}_{s}$ and $\mathbb{T}_t$, respectively.

For all experiments, we use the same underlying \texttt{transformer-iwslt} architecture in \texttt{fairseq}~\cite{ott-etal-2019-fairseq}, and only vary the embedding and decoding layers of the model by changing the tokenizer.
We use weight tying between the decoder side embedding and output weights.
The internal language model, without the embedding and decoding layers, has 31.5M parameters.\footnote{Thus, the total percentage of parameters contributed by the embedding and decoding layers can be computed as $\frac{(\hat{\mathbb{B}}_s + \hat{\mathbb{B}}_t)\times d}{31.5M + (\hat{\mathbb{B}}_s + \hat{\mathbb{B}}_t) \times d} \times 100$, where the embedding dimension~$d = 512$, unless otherwise noted.}
Complete model and training information is given in Table \ref{tbl:arch} in \autoref{apx:architecture}.
Via grid search, we found that the vocabulary size $(\mathbb{B}_s, \mathbb{B}_t) = (6k, 6k)$ performed the best, and use it as our optimal baseline.

For each hyperparameter setting, we report the BLEU score of the baseline and $\Delta$BLEU for its trimmed models,\footnote{We report the average of three training runs initialized with different random seeds. We note that using other metrics such as ChrF does not change the general trend of our findings.}
the \textit{effective vocabulary} size $(\hat{\mathbb{B}}_{s}, \hat{\mathbb{B}}_{t})$, which is the size of the resulting vocabularies after trimming with the given thresholds, and \textit{sequence length}, the average number of tokens in the tokenized source and target test corpora for the baseline and the percent relative difference for the trimmed models. 

In all tables, unless otherwise noted, the best performing trimmed model for a given baseline is \underline{underlined}, and the worst performing trimmed model is \doubleunderline{double underlined}.

\subsection{Trimming The Optimal Baseline} \label{ssec:optimalbaselinetrimming}
We first investigate whether or not trimming can improve the performance of the optimal baseline.
In \autoref{tab:optimalbaseline}, in all but one case ($\mathbb{T}_{s,t} = (100, 50)$) the BLEU score is lower than the baseline.
A second case ($\mathbb{T}_{s,t} = (50, 50)$) is close to the baseline, but this is likely due to the minor actual change in tokenization over the data.
In most cases, the trimmed models exhibit a 0.2--0.3 BLEU reduction, up to 0.46 in the worst case, where $\mathbb{T}_{s,t} = (200, 150)$.

\begin{table}
    \centering
    \resizebox{\columnwidth}{!}{%
    \begin{tabular}{cc|ccc} \toprule
    \makecell{Vocabulary \\ $(\mathbb{B}_s, \mathbb{B}_t)$} & \makecell{Thresholds \\ $(\mathbb{T}_s, \mathbb{T}_t)$} & BLEU & \makecell{Effective \\ Vocabulary \\ $(\hat{\mathbb{B}}_{s}, \hat{\mathbb{B}}_{t})$}  & \makecell{Sequence \\ Length \\ source/target} \\ \midrule
    \multirow{17}{*}{$(6k, 6k)$} & Baseline & $34.05$                      &      $(6.0k, 6.0k)$     &        $23.30$/$22.12$                              \\ \
    & $(50, 50)$                            & $-0.09$                      &     $(5.8k, 5.5k)$     &        $+0.2\%$/$+0.4\%$                              \\
    & $(50, 100)$                           & $-0.29$                      &     $(5.8k, 4.2k)$     &        $+0.2\%$/$+4.2\%$                              \\
    & $(50, 150)$                           & $-0.26$                      &     $(5.8k, 2.9k)$     &        $+0.2\%$/$+10.6\%$                              \\
    & $(50, 200)$                           & $+0.01$                      &     $(5.8k, 2.3k)$     &        $+0.2\%$/$+15.6\%$                              \\
    & $(100, 50)$                           & \underline{$+0.07$}          &     $(5.3k, 5.5k)$     &        $+1.4\%$/$+0.4\%$                              \\
    & $(100, 100)$                          & $-0.28$                      &     $(5.3k, 4.2k)$     &        $+1.4\%$/$+4.2\%$                              \\
    & $(100, 150)$                          & \doubleunderline{$-0.66$}    &     $(5.3k, 2.9k)$     &        $+1.4\%$/$+10.6\%$                              \\
    & $(100, 200)$                          & $-0.41$                      &     $(5.3k, 2.3k)$     &        $+1.4\%$/$+15.6\%$                              \\
    & $(150, 50)$                           & $-0.22$                      &     $(3.7k, 5.5k)$     &        $+7.5\%$/$+0.4\%$                              \\
    & $(150, 100)$                          & $-0.27$                      &     $(3.7k, 4.2k)$     &        $+7.5\%$/$+4.2\%$                              \\
    & $(150, 150)$                          & $-0.28$                      &     $(3.7k, 2.9k)$     &        $+7.5\%$/$+10.6\%$                              \\
    & $(150, 200)$                          & $-0.22$                      &     $(3.7k, 2.3k)$     &        $+7.5\%$/$+15.6\%$                              \\
    & $(200, 50)$                           & $-0.19$                      &     $(2.9k, 5.5k)$     &        $+13.1\%$/$+0.4\%$                              \\
    & $(200, 100)$                          & $-0.22$                      &     $(2.9k, 4.2k)$     &        $+13.1\%$/$+4.2\%$                              \\
    & $(200, 150)$                          & $-0.12$                      &     $(2.9k, 2.9k)$     &        $+13.1\%$/$+10.6\%$                              \\
    & $(200, 200)$                          & $-0.30$                      &     $(2.9k, 2.3k)$     &        $+13.1\%$/$+15.6\%$                              \\ \bottomrule                                   
    \end{tabular}}
    \caption{A comparison between the optimal baseline BPE model and its trimmed counterparts. \label{tab:optimalbaseline}}
\end{table}

\begin{table}[t]
    \centering
    \resizebox{\columnwidth}{!}{%
    \begin{tabular}{cc|ccc} \toprule
    \makecell{Vocabulary \\ $(\mathbb{B}_s, \mathbb{B}_t)$} & \makecell{Thresholds \\ $(\mathbb{T}_s, \mathbb{T}_t)$} & BLEU & \makecell{Effective \\ Vocabulary \\ $(\hat{\mathbb{B}}_{s}, \hat{\mathbb{B}}_{t})$} & \makecell{Sequence \\ Length \\ source/target} \\ \midrule
    \multirow{10}{*}{$(8k, 8k)$}  & Baseline & $33.63$                             &      $(8k, 8k)$               &        $22.47$/$21.51$                                   \\ 
                                      & $(100, 100)$ & $+0.16$                     &      $(4.8k, 3.7k)$           &        $+7.3\%$/$+9.4\%$                                   \\
                                      & $(100, 150)$ & $-0.02$                     &      $(4.8k, 2.6k)$           &        $+7.3\%$/$+16.7\%$                                   \\
                                      & $(100, 200)$ & \underline{$+0.32$}         &      $(4.8k, 2.1k)$           &        $+7.3\%$/$+22.0\%$                                  \\
                                      & $(150, 100)$ & $+0.24$                     &      $(3.3k, 3.7k)$           &        $+14.7\%$/$+9.4\%$                                   \\
                                      & $(150, 150)$ & $-0.01$                     &      $(3.3k, 2.6k)$           &        $+14.7\%$/$+16.7\%$                                   \\
                                      & $(150, 200)$ & $+0.05$                     &      $(3.3k, 2.1k)$           &        $+14.7\%$/$+22.0\%$                                   \\
                                      & $(200, 100)$ & $+0.27$                     &      $(2.6k, 3.7k)$           &        $+21.3\%$/$+9.4\%$                                   \\
                                      & $(200, 150)$ & \doubleunderline{$-0.03$}   &      $(2.6k, 2.6k)$           &        $+21.3\%$/$+16.7\%$                 \\
                                      & $(200, 200)$ & $+0.18$                     &      $(2.6k, 2.1k)$           &        $+21.3\%$/$+22.0\%$                                  \\ \midrule
    \multirow{10}{*}{$(10k, 10k)$} & Baseline & \doubleunderline{$33.56$}   &      $(10k, 9.9k)$               &        $21.93$/$21.12$                                   \\ 
                                        & $(100, 100)$ & \underline{$+0.37$}   &      $(4.4k, 3.4k)$           &        $+12.3\%$/$+13.2\%$                      \\
                                        & $(100, 150)$ & $+0.30$   &      $(4.4k, 2.4k)$                       &        $+12.3\%$/$+20.9\%$                                   \\
                                        & $(100, 200)$ & $+0.14$   &      $(4.4k, 1.9k)$                       &        $+12.3\%$/$+26.6\%$                                   \\
                                        & $(150, 100)$ & $+0.14$   &      $(3.1k, 3.4k)$                       &        $+20.1\%$/$+13.2\%$                                   \\
                                        & $(150, 150)$ & $+0.23$   &      $(3.1k, 2.4k)$                       &        $+20.1\%$/$+20.9\%$                                   \\
                                        & $(150, 200)$ & $+0.24$   &      $(3.1k, 1.9k)$                       &        $+20.1\%$/$+26.6\%$                                   \\
                                        & $(200, 100)$ & $+0.31$   &      $(2.3k, 3.4k)$                       &        $+27.3\%$/$+13.2\%$                                   \\
                                        & $(200, 150)$ & $+0.18$   &      $(2.3k, 2.4k)$                       &        $+27.3\%$/$+20.9\%$                                   \\
                                        & $(200, 200)$ & $+0.18$   &      $(2.3k, 1.9k)$                       &        $+27.3\%$/$+26.6\%$                                   \\ \midrule
    \multirow{10}{*}{$(12k, 12k)$}      & Baseline & $33.68$                         &    $(12k, 11.8k)$             &            $21.55$/$20.83$    \\ 
                                        & $(100, 100)$ & $+0.15$                     &      $(4.1k, 3.2k)$           &        $+15.9\%$/$+16.2\%$                                   \\
                                        & $(100, 150)$ & $+0.09$                     &      $(4.1k, 2.2k)$           &        $+15.9\%$/$+24.3\%$                                   \\
                                        & $(100, 200)$ & $-0.07$                     &      $(4.1k, 1.8k)$           &        $+15.9\%$/$+31.0\%$                                   \\
                                        & $(150, 100)$ & \underline{$+0.40$}         &      $(2.9k, 3.2k)$           &        $+24.7\%$/$+16.2\%$                       \\
                                        & $(150, 150)$ & \doubleunderline{$-0.16$}   &      $(2.9k, 2.2k)$           &        $+24.7\%$/$+24.3\%$                 \\
                                        & $(150, 200)$ & $-0.01$                     &      $(2.9k, 1.8k)$           &        $+24.7\%$/$+31.0\%$                                   \\
                                        & $(200, 100)$ & $+0.28$                     &      $(2.2k, 3.2k)$           &        $+32.4\%$/$+16.2\%$                                   \\
                                        & $(200, 150)$ & $+0.11$                     &      $(2.2k, 2.2k)$           &        $+32.4\%$/$+24.3\%$                                   \\
                                        & $(200, 200)$ & $+0.16$                     &      $(2.2k, 1.8k)$           &        $+32.4\%$/$+31.0\%$                                   \\ \bottomrule
    \end{tabular}}
    \caption{A comparison between several suboptimal baselines and their trimmed counterparts.}  \label{tab:suboptimalbaseline} %
\end{table}

\subsection{Trimming Suboptimal Baselines} \label{ssec:suboptimalbaselinetrimming}
In Section \ref{ssec:optimalbaselinetrimming}, we found that trimming the optimal baseline had a slight negative effect.
However, it is still possible that \textit{suboptimal} baselines have some inherent issue that could be resolved by trimming.
We believe that practitioners are generally interested in heuristics to improve their models \textit{without} resorting to huge hyperparameter sweeps, and so
we consider the much more likely situation where we begin at a suboptimal BPE configuration.

In \autoref{tab:suboptimalbaseline}, we present results for several baseline configurations that underperform our optimal baseline, along with various trimming thresholds. %
In the $\mathbb{B}_{s,t} = (10k, 10k)$ case, the worst-performing baseline, we see that every trimmed model improves upon the baseline, at most by $+$0.37 BLEU.
However, in the other two cases, this effect largely goes away, with the trimmed BLEU scores being more centered around the baseline.
Thus, it appears that trimming may help recover some performance in very-low-performing models, but this does not reflect a consistently positive trend.

\begin{table*}[t]
    \centering
    \resizebox{1.9\columnwidth}{!}{%
    \begin{tabular}{cc|ccc} \toprule
    \makecell{Vocabulary \\ $(\mathbb{B}_s, \mathbb{B}_t)$} & \makecell{Thresholds \\ $(\mathbb{T}_s, \mathbb{T}_t)$} & BLEU & \makecell{Effective \\ Vocabulary \\ $(\hat{\mathbb{B}}_{s}, \hat{\mathbb{B}}_{t})$} & \makecell{Sequence \\ Length \\ source/target} \\ \midrule
    \multirow{6}{*}{$(9k, 12k)$} & Baseline & $33.74$                           &      $(9k, 12k)$            &        $22.17$/$20.83$                                   \\ \
                                    & $(0, 100)$  & $-0.2$                      &      $(9k, 3.2k)$           &        $~~~~-$/$+16.2\%$                                   \\
                                    & $(0, 150)$  & \underline{$+0.21$}         &      $(9k, 2.2k)$           &        $~~~~-$/$+24.3\%$                                   \\
                                    & $(0, 200)$  & $+0.06$                     &      $(9k, 1.8k)$           &        $~~~~-$/$+31.0\%$                                   \\
                                    & $(0, 250)$  & $+0.01$                     &      $(9k, 1.5k)$           &        $~~~~-$/$+38.0\%$                                   \\
                                    & $(0, 300)$  & \doubleunderline{$-0.27$}   &      $(9k, 1.3k)$           &        $~~~~-$/$+44.6\%$                                   \\ \midrule
    \multirow{11}{*}{$(6k, 6k)$} & Baseline   & $34.05$                         &      $(6k, 6k)$             &        $23.30$/$22.12$                                   \\ \
                                    & $(100, 0)$  & \underline{$+0.05$}         &      $(5.3k, 6k)$           &        $+1.4\%$/$-~~~~~$                                   \\
                                    & $(150, 0)$  & $-0.23$                     &      $(3.7k, 6k)$           &        $+7.5\%$/$-~~~~~$                                   \\
                                    & $(200, 0)$  & $-0.13$                     &      $(2.9k, 6k)$           &        $+13.1\%$/$-~~~~~~~$                                   \\
                                    & $(250, 0)$  & $-0.24$                     &      $(2.3k, 6k)$           &        $+18.5\%$/$-~~~~~~~$                                   \\
                                    & $(300, 0)$  & $-0.24$                     &      $(2.0k, 6k)$           &        $+23.2\%$/$-~~~~~~~$                                   \\
                                    & $(0, 100)$  & $-0.17$                     &      $(6k, 4.2k)$           &        $~~~~-$/$+4.2\%$                                   \\
                                    & $(0, 150)$  & $-0.07$                     &      $(6k, 2.9k)$           &        $~~~~~-$/$+10.6\%$                                   \\
                                    & $(0, 200)$  & $-0.08$                     &      $(6k, 2.3k)$           &        $~~~~~-$/$+15.6\%$                                   \\
                                    & $(0, 250)$  & \doubleunderline{$-0.59$}   &      $(6k, 1.9k)$           &        $~~~~~-$/$+20.3\%$                 \\
                                    & $(0, 300)$  & $-0.20$                     &      $(6k, 1.6k)$           &        $~~~~~-$/$+24.2\%$                                   \\ \bottomrule
    \end{tabular}
    \hspace{0.5cm}
    \begin{tabular}{cc|ccc} \toprule
    \makecell{Vocabulary \\ $(\mathbb{B}_s, \mathbb{B}_t)$} & \makecell{Thresholds \\ $(\mathbb{T}_s, \mathbb{T}_t)$} & BLEU & \makecell{Effective \\ Vocabulary \\ $(\hat{\mathbb{B}}_{s}, \hat{\mathbb{B}}_{t})$} & \makecell{Sequence \\ Length \\ source/target} \\ \midrule
    \multirow{6}{*}{$(12k, 9k)$} & Baseline & $33.76$                           &      $(12k, 8.9k)$            &        $21.55$/$21.30$                                   \\ \
                                    & $(100, 0)$  & $-0.03$                     &      $(4.1k, 8.9k)$           &        $+15.9\%$/$-~~~~~~$                                   \\
                                    & $(150, 0)$  & \underline{$+0.03$}         &      $(2.9k, 8.9k)$           &        $+24.7\%$/$-~~~~~~$                                   \\
                                    & $(200, 0)$  & $-0.12$                     &      $(2.2k, 8.9k)$           &        $+32.4\%$/$-~~~~~~$                                   \\
                                    & $(250, 0)$  & $-0.23$                     &      $(1.8k, 8.9k)$           &        $+39.4\%$/$-~~~~~~$                                   \\
                                    & $(300, 0)$  & \doubleunderline{$-0.26$}   &      $(1.5k, 8.9k)$           &        $+46.5\%$/$-~~~~~~$                                   \\ \midrule
    \multirow{11}{*}{$(10k, 10k)$} & Baseline & $33.56$                         &      $(10k, 9.9k)$            &        $21.93$/$21.12$                                   \\ \
                                    & $(100, 0)$  & $+0.02$                     &      $(4.4k, 9.9k)$           &        $+12.3\%$/$-~~~~~~$                                   \\
                                    & $(150, 0)$  & $-0.09$                     &      $(3.1k, 9.9k)$           &        $+20.1\%$/$-~~~~~~$                                   \\
                                    & $(200, 0)$  & $-0.06$                     &      $(2.3k, 9.9k)$           &        $+27.3\%$/$-~~~~~~$                                   \\
                                    & $(250, 0)$  & $-0.18$                     &      $(1.9k, 9.9k)$           &        $+33.5\%$/$-~~~~~~$                                   \\
                                    & $(300, 0)$  & \doubleunderline{$-0.21$}   &      $(1.6k, 9.9k)$           &        $+40.6\%$/$-~~~~~~$                  \\
                                    & $(0, 100)$  & $-0.12$                     &      $(10k, 3.4k)$            &        $~~~~-$/$+13.2\%$                                   \\
                                    & $(0, 150)$  & \underline{$+0.27$}         &      $(10k, 2.4k)$            &        $~~~~-$/$+20.9\%$                                   \\
                                    & $(0, 200)$  & $+0.10$                     &      $(10k, 1.9k)$            &        $~~~~-$/$+26.6\%$                                   \\
                                    & $(0, 250)$  & $+0.12$                     &      $(10k, 1.6k)$            &        $~~~~-$/$+33.0\%$                                   \\
                                    & $(0, 300)$  & $-0.09$                     &      $(10k, 1.3k)$            &        $~~~~-$/$+38.7\%$                                   \\ \bottomrule         
    \end{tabular}}
    \caption{Results for trimming only the source language ($\mathbb{T}_t = 0$) or only the target language ($\mathbb{T}_s = 0$). \label{tab:srcvstgt}}
\end{table*}

\subsection{The Effect of Trimming Source vs Target} \label{ssec:sourcevstarget}
Since trimming is applied independently for the source and target languages, it is possible that trimming one more aggressively than the other would have downstream effects on the model.
On one hand, it may be that trimming the source more aggressively would produce a better model, in that a reduction in very-rare source subwords can lead to more coherent contexts.
This is not a problem for the target side, as under the BLEU or chrF metrics, the model has no requirement to \say{spell} an output using any particular combination of subwords. 
On the other hand, one can argue that trimming the target vocabulary can simplify the generation side's softmax operation, leading to better training.

In Table \ref{tab:srcvstgt}, we compare a set of baselines to models where \textit{only} the source or \textit{only} the target is trimmed.
In two of our examples $(\mathbb{B}_s, \mathbb{B}_t) = (9k, 12k)$ and $(12k, 9k)$), we choose BPE configurations where either the source or the target has a much larger base vocabulary, and then only trim that side.
Like our previous findings, we find that trimming in this way does not improve upon either optimal or suboptimal baselines in a meaningful way.
However, we also observe that there is a consistent negative trend when trimming too much from the source vocabulary. The same is not observed in other experiments when holding the target side threshold constant and increasing the source side.
For example, in \autoref{tab:suboptimalbaseline}, no such trend is observed between the $\mathbb{T}_s \in \{100, 150, 200\}$ and $\mathbb{T}_t = 100$ hyperparameter settings for any baseline.
One possible explanation is that when trimming only one side, the vocabulary sizes become so mismatched that the model cannot properly represent even surface-level mappings, but then we would expect for this effect to also appear when aggressively trimming the target side only, or have a lesser effect when starting with a mismatched vocabulary size, both of which are not observed empirically.
We thus conclude that trimming only one side has no consistent positive effect on model quality, and that trimming the source language too aggressively in isolation has an overall negative effect.

\subsection{Trimming Such That 95\% of Tokens Appear More Than 100 Times} \label{ssec:100heuristic}
Finding the optimal vocabulary size is a challenging issue. \citet{gowda-may-2020-finding} perform extensive hyperparameter tests and arrive at the following heuristic: \textit{pick the largest vocabulary such that $>$95\% of tokens appear more than 100 times}.
We approximate this by simply setting $\mathbb{T}_{s} = \mathbb{T}_t = 100$, which is a slightly different setting than their suggestion. %
We first note that the $\mathbb{B}_{s,t} = (6k, 6k)$ optimal baseline \textit{does not} have this property.
Only 88\% of the source tokens and 70\% of the target tokens (79\% overall) appear more than 100 times. 

In \autoref{tbl:100thresholdexperiment}, we compare several baselines to their counterparts with $\mathbb{T}_{s} = \mathbb{T}_t = 100$.
In 31 out of 36 cases, the trimmed model was within $\pm$0.2 BLEU of its baseline.
The largest positive improvement was when $\mathbb{B}_{s,t} = (10k, 10k)$, with a maximum increase of 0.37 BLEU. %
This suggests that trimming such that all tokens have frequency at least 100 has, at best, only a slight positive effect for suboptimal BPE configurations.

\begin{table*}
    \centering
    \resizebox{1.5\columnwidth}{!}{%
        \begin{tabular}{c|cccccc} \toprule 
                                    
            \backslashbox{$\mathbb{B}_s$}{$\mathbb{B}_t$} & $5k$  & $6k$ & $7k$ & $8k$ & $9k$ & $10k$ \\ \midrule
            $5k$   & $33.88$/$\bf{33.89}$ & $\bf{33.91}$/$33.90$ & $33.79$/$33.92$ & $33.80$/$\bf{33.83}$ & $33.65$/$\bf{33.81}$ & $33.73$/$\bf{33.86}$  \\
            $6k$   & $33.74$/$\bf{33.87}$ & $\bf{34.05}$/$33.77$ & $\bf{33.87}$/$33.85$ & $\bf{33.81}$/$\bf{33.81}$ & $\bf{33.89}$/$33.81$ & $33.79$/$\bf{33.89}$  \\
            $7k$   & $33.83$/$\bf{33.97}$ & $\bf{33.85}$/$\bf{33.85}$ & $\bf{33.85}$/$33.76$ & $33.47$/$\bf{33.83}$ & $33.93$/$\bf{33.97}$ & $\bf{33.83}$/$33.78$  \\
            $8k$   & $\bf{33.95}$/$33.75$ & $33.92$/$\bf{34.03}$ & $33.74$/$\bf{33.83}$ & $33.63$/$\bf{33.79}$ & $\bf{33.90}$/$33.86$ & $33.89$/$\bf{34.04}$  \\
            $9k$   & $33.84$/$\bf{33.95}$ & $33.94$/$\bf{34.10}$ & $\bf{33.89}$/$33.80$ & $33.66$/$\bf{33.86}$ & $33.74$/$\bf{34.00}$ & $33.68$/$\bf{33.92}$  \\
            $10k$  & $\bf{33.85}$/$33.82$ & $33.87$/$\bf{33.89}$ & $\bf{33.80}$/$33.78$ & $33.74$/$\bf{34.01}$ & $33.86$/$\bf{33.92}$ & $33.56$/$\bf{33.93}$  \\ \bottomrule
        \end{tabular}}
        
        \caption{A comparison between baseline split-vocabulary BPE models and the same model but trimmed with $\mathbb{T}_{s} = \mathbb{T}_t = 100$. Each cell contains the BLEU score for the baseline followed by the BLEU score for the trimmed model, and the better-performing model is \bf{bolded}.\label{tbl:100thresholdexperiment}}

\end{table*}

\subsection{The Effect of Trimming on Rare-Subword Sentences} \label{ssec:raresentences}
We have generally seen that trimming has little positive effect on the downstream quality of the model. However, we are trimming only rare subwords, which by definition appear only in a few sentences each.
Perhaps, even if the overall model quality does not improve, the translation of sentences that include rare subwords could improve after trimming, as rare subwords are replaced by more common subwords with more robust embeddings.
This in itself would be a valid \say{selling point} for trimming, as a means of avoiding data-induced errors and for robustness in low-resource settings.

In this setting, given a baseline tokenizer, we select a threshold $\mathbb{T}$.
Then, for both the source and target side sentences, we select the subset of sentences in the testing corpus that, after being tokenized with the baseline tokenizer, contain a subword that would have been trimmed from the model if a threshold $\mathbb{T}$ had been applied.
As \autoref{tab:raresentences} in \autoref{app:rare}\footnote{Moved to Appendix \ref{app:rare} due to the table's size.} shows, no patterns emerge along any of the settings we control for.

\subsection{The Effect of Preserving Terminal Subwords} \label{ssec:terminalpreservation}
In our trimming process, we do not differentiate between trimming \emph{intermediate} subwords (that can participate in larger merges) and \emph{terminal} subwords (that do not form part of a larger merge).
In many cases, terminal subwords can represent full words or concepts, even if rare, and perhaps it is beneficial to not trim them.
Terminal subwords also have the property that their frequency in the corpus is maximal, in the sense that they were added to the vocabulary due to having the highest frequency at the time, and having never been subsumed into another token, kept their frequency.

\begin{table*}[t]
    \centering
    \resizebox{1.5\columnwidth}{!}{%
    \begin{tabular}{cc|ccc} \toprule
    \makecell{Vocabulary \\ $(\mathbb{B}_s, \mathbb{B}_t)$} & \makecell{Thresholds \\ $(\mathbb{T}_s, \mathbb{T}_t)$} & \makecell{BLEU \\ trimmed, preserving} & \makecell{Effective Vocabulary \\ trimmed / preserving \\ $(\hat{\mathbb{B}}_{s}, \hat{\mathbb{B}}_{t})$ / $(\hat{\mathbb{B}}_{s}, \hat{\mathbb{B}}_{t})$} & \makecell{Sequence Length \\ (trimmed, preserving) \\ source/target} \\ \midrule
    \multirow{10}{*}{$(6k, 6k)$} & Baseline & \underline{$34.05$}                           &             $(6.1k, 6.0k)$                     &               $23.30$/$22.12$                                   \\ \
                                    & $(100, 100)$  & $-0.28$, $-0.19$                      &      $(5.3k, 4.2k)$/$(5.5k, 5.0k)$           &        $+1.4\%$/$+4.2\%$, $+0.7\%$/$+1.4\%$            \\
                                    & $(100, 150)$  & \doubleunderline{$-0.66$}, $-0.19$    &      $(5.3k, 2.9k)$/$(5.5k, 4.8k)$           &        $+1.4\%$/$+10.6\%$, $+0.7\%$/$+2.4\%$            \\
                                    & $(100, 200)$  & $-0.41$, $-0.28$                      &      $(5.3k, 2.3k)$/$(5.5k, 4.6k)$           &        $+1.4\%$/$+15.6\%$, $+0.7\%$/$+3.5\%$            \\
                                    & $(150, 100)$  & $-0.27$, $-0.15$                      &      $(3.7k, 4.2k)$/$(5.2k, 5.0k)$           &        $+7.5\%$/$+4.2\%$, $+1.7\%$/$+1.4\%$            \\
                                    & $(150, 150)$  & $-0.28$, $-0.30$                      &      $(3.7k, 2.9k)$/$(5.2k, 4.8k)$           &        $+7.5\%$/$+10.6\%$, $+1.7\%$/$+2.4\%$            \\
                                    & $(150, 200)$  & $-0.22$, $-0.31$                      &      $(3.7k, 2.3k)$/$(5.2k, 4.6k)$           &        $+7.5\%$/$+15.6\%$, $+1.7\%$/$+3.5\%$            \\
                                    & $(200, 100)$  & $-0.22$, $-0.09$                      &      $(2.9k, 4.2k)$/$(5.0k, 5.0k)$           &        $+13.1\%$/$+4.2\%$, $+2.9\%$/$+1.4\%$            \\
                                    & $(200, 150)$  & $-0.12$, $-0.14$                      &      $(2.9k, 2.9k)$/$(5.0k, 4.8k)$           &        $+13.1\%$/$+10.6\%$, $+2.9\%$/$+2.4\%$            \\
                                    & $(200, 200)$  & $-0.30$, $-0.27$                      &      $(2.9k, 2.3k)$/$(5.0k, 4.6k)$           &        $+13.1\%$/$+15.6\%$, $+2.9\%$/$+3.5\%$            \\ \bottomrule

    \multirow{10}{*}{$(9k, 9k)$} & Baseline         &      $33.74$                          &             $(9.0k, 8.9k)$                   &                $22.17$/$21.30$                                   \\ \
                                    & $(100, 100)$  & $+0.26$, \underline{$+0.30$}          &      $(4.6k, 3.6k)$/$(7.6k, 6.9k)$           &        $+9.9\%$/$+11.5\%$, $+2.2\%$/$+2.6\%$            \\
                                    & $(100, 150)$  & $+0.03$, $-0.03$                      &      $(4.6k, 2.5k)$/$(7.6k, 6.6k)$           &        $+9.9\%$/$+19.0\%$, $+2.2\%$/$+3.9\%$            \\
                                    & $(100, 200)$  & $+0.13$, $+0.18$                      &      $(4.6k, 2.0k)$/$(7.6k, 6.4k)$           &        $+9.9\%$/$+24.6\%$, $+2.2\%$/$+5.0\%$            \\
                                    & $(150, 100)$  & \doubleunderline{$-0.10$}, $+0.10$    &      $(3.2k, 3.6k)$/$(7.2k, 6.9k)$           &        $+17.8\%$/$+11.5\%$, $+4.1\%$/$+2.6\%$            \\
                                    & $(150, 150)$  & $+0.24$, $+0.03$                      &      $(3.2k, 2.5k)$/$(7.2k, 6.6k)$           &        $+17.8\%$/$+19.0\%$, $+4.1\%$/$+3.9\%$            \\
                                    & $(150, 200)$  & $+0.11$, $-0.01$                      &      $(3.2k, 2.0k)$/$(7.2k, 6.4k)$           &        $+17.8\%$/$+24.6\%$, $+4.1\%$/$+5.0\%$            \\
                                    & $(200, 100)$  & $+0.08$, $+0.07$                      &      $(2.5k, 3.6k)$/$(6.9k, 6.9k)$           &        $+24.4\%$/$+11.5\%$, $+5.6\%$/$+2.6\%$            \\
                                    & $(200, 150)$  & $+0.05$, $+0.11$                      &      $(2.5k, 2.5k)$/$(6.9k, 6.6k)$           &        $+24.4\%$/$+19.0\%$, $+5.6\%$/$+3.9\%$            \\
                                    & $(200, 200)$  & $+0.01$, $+0.05$                      &      $(2.5k, 2.0k)$/$(6.9k, 6.4k)$           &        $+24.4\%$/$+24.6\%$, $+5.6\%$/$+5.0\%$            \\ \bottomrule            
    \end{tabular}}
    \caption{A comparison of trimmed BPE models without and with terminal-subword preservation. Each cell contains the values for the trimmed tokenizer and the terminal-preserving trimmed tokenizer, separated by a comma. \label{tab:terminalpreservation}}
\end{table*}

In \autoref{tab:terminalpreservation}, we select a number of baselines' trimming hyperparameters, and compare the effect of preserving vs removing terminal subwords.
In all but two cases, the difference in score between the preserved-terminal models and the regular trimmed models is within $\pm$0.2 BLEU.
For larger thresholds, preserving terminals massively reduces the number of tokens that are trimmed, thus we should not expect the performance of the models or the model size to change much as we increase the threshold while preserving terminals.
Overall, we observe no consistent trend between preserving or not preserving terminal subwords and the baseline.

\subsection{Trimming vs Initializing Smaller Vocabularies} \label{ssec:stunted}
Trimming a BPE tokenizer reduces the effective vocabulary size, which could lead to an argument that models with trimmed tokenizers should be compared to untrimmed models of similar effectively vocabulary sizes.
In \autoref{tab:stunted}, we compare trimmed models to untrimmed BPE models that are initialized to have the same effective vocabulary size.

In most cases (six out of nine of our configurations), the smaller-initialized model outperforms the same-effective-size trimmed model.
However, one benefit that the trimmed model might have is that the final tokenized sequences can be shorter, if the trimming removes short, intermediate subwords.
However, we see that in nearly every case (and every case where $\mathbb{T} > 100$), the untrimmed models produce shorter sequences on average.
This indicates that, given the same parameter budget for the tokenizer, the naive BPE initialization is a better choice than initializing a larger vocabulary and trimming to the desired size.

\begin{table}[t]
    \centering
    \resizebox{\columnwidth}{!}{%
    \begin{tabular}{cc|ccc} \toprule
    \makecell{Vocabulary \\ $(\mathbb{B}_s, \mathbb{B}_t)$} & \makecell{Thresholds \\ $(\mathbb{T}_s, \mathbb{T}_t)$} & BLEU & \makecell{Effective \\ Vocabulary \\ $(\hat{\mathbb{B}}_{s}, \hat{\mathbb{B}}_{t})$} & \makecell{Sequence \\ Length \\ source/target} \\ \midrule
    \multirow{7}{*}{$(6k, 6k)$}       & Baseline             & \underline{$34.05$} &      $(6.1k, 6.0k)$             &        $23.30$/$22.12$      \\ \
                                      & Trimmed $(100, 100)$ & $-0.28$   &      $(5.3k, 4.2k)$           &        $+1.4\%$/$+4.2\%$      \\ \
                                      & Untrimmed            & \doubleunderline{$-0.32$}   &      $(5.3k, 4.2k)$           &        $+1.8\%$/$+4.9\%$      \\ \
                                      & Trimmed $(150, 150)$ & $-0.28$   &      $(3.7k, 2.9k)$           &        $+7.5\%$/$+10.6\%$      \\ \
                                      & Untrimmed            & $-0.10$   &      $(3.8k, 2.9k)$           &        $+7.2\%$/$+10.6\%$      \\ \
                                      & Trimmed $(200, 200)$ & $-0.30$   &      $(2.9k, 2.3k)$           &        $+13.1\%$/$+15.6\%$      \\ \
                                      & Untrimmed            & $-0.10$   &      $(2.9k, 2.3k)$           &        $+11.9\%$/$+15.3\%$      \\ \midrule
    \multirow{7}{*}{$(8k, 8k)$}       & Baseline             & $33.63$   &      $(8k, 8k)$               &        $22.47$/$21.51$      \\ \
                                      & Trimmed $(100, 100)$ & $+0.16$   &      $(4.8k, 3.7k)$           &        $+7.3\%$/$+9.4\%$      \\ \
                                      & Untrimmed            & \underline{$+0.31$}   &      $(4.9k, 3.8k)$           &        $+6.9\%$/$+9.6\%$      \\ \
                                      & Trimmed $(150, 150)$ & $-0.01$   &      $(3.3k, 2.6k)$           &        $+14.7\%$/$+16.7\%$      \\ \
                                      & Untrimmed            & \doubleunderline{$-0.16$}   &      $(3.4k, 2.6k)$           &        $+13.2\%$/$+16.1\%$      \\ \
                                      & Trimmed $(200, 200)$ & $+0.18$   &      $(2.6k, 2.1k)$           &        $+21.3\%$/$+22.0\%$      \\ \
                                      & Untrimmed            & $+0.23$   &      $(2.6k, 2.1k)$           &        $+18.7\%$/$+20.8\%$      \\ \midrule
    \multirow{7}{*}{$(10k, 10k)$} & Baseline                 & \doubleunderline{$33.56$}   &      $(10k, 9.9k)$            &        $21.93$/$21.12$      \\ \
                                      & Trimmed $(100, 100)$ & $+0.37$   &      $(4.4k, 3.4k)$           &        $+12.3\%$/$+13.2\%$      \\ \
                                      & Untrimmed            & $+0.37$   &      $(4.5k, 3.5k)$           &        $+11.1\%$/$+12.9\%$      \\ \
                                      & Trimmed $(150, 150)$ & $+0.23$   &      $(3.1k, 2.4k)$           &        $+20.1\%$/$+20.9\%$      \\ \
                                      & Untrimmed            & \underline{$+0.42$}   &      $(3.1k, 2.4k)$           &        $+17.6\%$/$+19.8\%$      \\ \
                                      & Trimmed $(200, 200)$ & $+0.18$   &      $(2.3k, 1.9k)$           &        $+27.3\%$/$+26.6\%$      \\ \
                                      & Untrimmed            & $+0.34$   &      $(2.4k, 2.0k)$           &        $+23.8\%$/$+24.6\%$      \\ \bottomrule         
    \end{tabular}}
    \caption{A comparison between baseline models with a given $(\mathbb{B}_s, \mathbb{B}_t)$ and larger models that are trimmed to the same effective vocabulary size. \label{tab:stunted}}
\end{table}

\subsection{Joint Vocab} \label{ssec:joint_vocab}

We now consider the joint vocabulary setting.
Here, we select a single BPE size parameter $\mathbb{B}_j$ and train the vocabulary on the concatenation of the source and target corpora.
By default, only tokens which appear at least once in each language's tokenized corpus are included in the embedding table, making the effective vocabulary size for even untrimmed models less than the BPE initialization settings.

\autoref{tab:joint_experiments} lists the experimental results for $\mathbb{B}_j = 7k, 10k, 14k$.
As in the split vocabulary setting, trimming generally reduces model performance in the joint setting.
In only one out of 30 configurations does the trimmed model outperform its baseline ($\mathbb{B}_j = 10k, \mathbb{T}_{s,t} = (100, 100)$), while in all other cases we observe a consistent drop.

\begin{table}[t]
    \centering
    \resizebox{\columnwidth}{!}{%
    \begin{tabular}{cc|ccc} \toprule
    \makecell{Vocabulary \\ $\mathbb{B}_j$} & \makecell{Thresholds \\ $(\mathbb{T}_s, \mathbb{T}_t)$} & BLEU & \makecell{Effective \\ Vocabulary \\ $(\hat{\mathbb{B}}_{s}, \hat{\mathbb{B}}_{t})$} & \makecell{Sequence \\ Length \\ source/target} \\ \midrule
    \multirow{10}{*}{$7k$} & Baseline & \underline{$34.02$}        &      $(6.5k, 4.9k)$           &        $24.11$/$23.25$                                   \\
                             & (100, 100)  & $-0.02$   &      $(4.2k, 3.7k)$           &        $+1.8\%$/$+1.1\%$                                   \\
                             & (100, 150)  & $-0.15$   &      $(4.2k, 3.3k)$           &        $+1.8\%$/$+2.6\%$                                   \\
                             & (100, 200)  & \doubleunderline{$-0.54$}   &      $(4.2k, 2.7k)$           &        $+1.8\%$/$+6.1\%$                                   \\
                             & (150, 100)  & $-0.26$   &      $(3.8k, 3.7k)$           &        $+3.2\%$/$+1.1\%$                                   \\
                             & (150, 150)  & $-0.19$   &      $(3.8k, 3.3k)$           &        $+3.2\%$/$+2.6\%$                                   \\
                             & (150, 200)  & $-0.45$   &      $(3.8k, 2.7k)$           &        $+3.2\%$/$+6.1\%$                                   \\
                             & (200, 100)  & $-0.09$   &      $(3.1k, 3.7k)$           &        $+6.9\%$/$+1.1\%$                                   \\
                             & (200, 150)  & $-0.09$   &      $(3.1k, 3.3k)$           &        $+6.9\%$/$+2.6\%$                                   \\
                             & (200, 200)  & \underline{$~~=~~$}       &      $(3.1k, 2.7k)$           &        $+6.9\%$/$+6.1\%$                                   \\ \midrule
                             
    \multirow{10}{*}{$10k$} & Baseline & $34.02$       &      $(8.8k, 6.6k)$           &        $22.99$/$22.25$                                             \\ 
                             & (100, 100)  & \underline{$+0.15$}   &      $(5.1k, 4.3k)$           &        $+3.4\%$/$+3.1\%$                                   \\
                             & (100, 150)  & $-0.10$   &      $(5.1k, 3.0k)$           &        $+3.4\%$/$+9.5\%$                                   \\
                             & (100, 200)  & $-0.17$   &      $(5.1k, 2.3k)$           &        $+3.4\%$/$+14.5\%$                                   \\
                             & (150, 100)  & $-0.17$   &      $(3.6k, 4.3k)$           &        $+10.2\%$/$+3.1\%$                                   \\
                             & (150, 150)  & $-0.20$   &      $(3.6k, 3.0k)$           &        $+10.2\%$/$+9.5\%$                                   \\
                             & (150, 200)  & \doubleunderline{$-0.23$}   &      $(3.6k, 2.3k)$           &        $+10.2\%$/$+14.5\%$                                   \\
                             & (200, 100)  & $-0.12$   &      $(2.8k, 4.3k)$           &        $+15.9\%$/$+3.1\%$                                   \\
                             & (200, 150)  & $-0.11$   &      $(2.8k, 3.0k)$           &        $+15.9\%$/$+9.5\%$                                   \\
                             & (200, 200)  & $-0.17$   &      $(2.8k, 2.3k)$           &        $+15.9\%$/$+14.5\%$                                   \\ \midrule

    \multirow{10}{*}{$14k$} & Baseline & \underline{$33.94$}         &      $(12k, 8.9k)$           &        $22.09$/$21.56$                                   \\
                             & $(100, 100)$  & $-0.39$   &      $(4.6k, 3.8k)$          &        $+10.4\%$/$+8.9\%$                                   \\
                             & $(100, 150)$  & $-0.20$   &      $(4.6k, 2.6k)$          &        $+10.4\%$/$+16.0\%$                                   \\
                             & $(100, 200)$  & $-0.30$   &      $(4.6k, 2.0k)$          &        $+10.4\%$/$+21.7\%$                                   \\
                             & $(150, 100)$  & $-0.07$   &      $(3.1k, 3.8k)$          &        $+18.7\%$/$+8.9\%$                                   \\
                             & $(150, 150)$  & \doubleunderline{$-0.44$}   &      $(3.1k, 2.6k)$          &        $+18.7\%$/$+16.0\%$                                   \\
                             & $(150, 200)$  & $-0.22$   &      $(3.1k, 2.0k)$          &        $+18.7\%$/$+21.7\%$                                   \\
                             & $(200, 100)$  & $-0.21$   &      $(2.4k, 3.8k)$          &        $+25.5\%$/$+8.9\%$                                   \\
                             & $(200, 150)$  & $-0.41$   &      $(2.4k, 2.6k)$          &        $+25.5\%$/$+16.0\%$                                   \\
                             & $(200, 200)$  & $-0.26$   &      $(2.4k, 2.0k)$          &        $+25.5\%$/$+21.7\%$                                   \\ \bottomrule
             
    \end{tabular}}
    \caption{Trimming in a joint vocabulary setting. \label{tab:joint_experiments}}
\end{table}

\subsection{Larger Dataset} \label{ssec:larger_dataset}
In all prior experiments, we worked on the IWSLT14 German\textrightarrow{}English dataset, which is relatively small, chosen specifically due to the large number of experiments performed.
To show that our results extend beyond this single dataset, we repeat part of our experiments on the Europarl English\textrightarrow{}French dataset, which consists of 2 million sentence pairs \cite{koehn-2005-europarl}.
We also used a slightly larger transformer model, correspondong to \texttt{transformer} from \texttt{fairseq} (see Table \ref{tab:large_arch} in \autoref{apx:architecture})~\cite{ott-etal-2019-fairseq}.
Since it was too costly to run an extensive hyperparameter sweep to find the optimal BPE baselines, we picked three reasonable baselines ($10k$/$10k$, $20k$/$20k$, $30k$/$30k$) and assume that these are all sub-optimal.
We chose $\mathbb{T} = (50, 50), (100, 100), (150, 150), (200, 200)$ and, for each hyperparameter setting, trained three models and averaged their BLEU scores.

The results are presented in \autoref{tab:suboptimalbaseline_french}.
For the $10k$/$10k$ and $20k$/$20k$ settings, we observe largely the same trend as before:
trimming does not appreciably improve performance, and for the most part also does not reduce it.
However, unlike nearly all other settings, in the $30k$/$30k$ case, we observe a large increase in BLEU as we increase the trimming threshold. 
Nevertheless, even the best $30k$/$30k$ performance does not reach that of the other sizes' baselines.
We conjecture that, due to the substantially larger corpus, the models are better able to learn robust embeddings for most subwords. Compare, for example, the $10k$/$10k$ baselines in this setting and in the original DE--EN setting (Section \ref{ssec:suboptimalbaselinetrimming},  \autoref{tab:suboptimalbaseline}).
Only $\sim$500 subwords appear fewer than 100 times in the larger corpus, compared with over 5,000 in the DE--EN corpus.

Nevertheless, the results on $30k$ vocabularies prompted us to take a closer look into large initial vocabularies, which we present in Section \ref{ssec:large}.

\begin{table}
    \centering
    \resizebox{\columnwidth}{!}{%
    \begin{tabular}{cc|ccc} \toprule
    \makecell{Vocabulary \\ $(\mathbb{B}_s, \mathbb{B}_t)$} & \makecell{Thresholds \\ $(\mathbb{T}_s, \mathbb{T}_t)$} & BLEU & \makecell{Effective \\ Vocabulary \\ $(\hat{\mathbb{B}}_{s}, \hat{\mathbb{B}}_{t})$} & \makecell{Sequence \\ Length \\ source/target} \\ \midrule
    \multirow{5}{*}{$(10k, 10k)$} & Baseline & $43.41$              &    $(10k, 10k)$               &        $30.40$/$33.72$                                  \\ 
                                        & $(50, 50)$ & $-0.06$      &      $(9.7k, 9.8k)$           &        ${>}0.1\%$/${>}0.1\%$                                   \\
                                        & $(100, 100)$ & \underline{$+0.03$}    &     $(9.4k, 9.6k)$            &        ${>}0.1\%$/${>}0.1\%$                                   \\
                                        & $(150, 150)$ & \doubleunderline{$-0.11$}    &      $(9.3k, 9.6k)$           &        $+0.1\%$/$+0.1\%$                                   \\
                                        & $(200, 200)$ & $-0.06$    &      $(9.1k, 9.4k)$           &        $+0.1\%$/$+0.1\%$                       \\  \midrule
    \multirow{5}{*}{$(20k, 20k)$} & Baseline & $43.18$              &    $(19.8k, 19.9k)$           &        $28.96$/$31.93$                                   \\ 
                                        & $(50, 50)$ & $-0.07$      &      $(17.6k, 18.5k)$         &        ${>}0.1\%$/${>}0.1\%$                                   \\
                                        & $(100, 100)$ & \doubleunderline{$-0.09$}    &      $(16.5k, 17.8k)$         &        $+0.2\%$/$+0.2\%$                                   \\
                                        & $(150, 150)$ & \underline{$+0.03$}    &      $(14.0k, 17.2k)$         &        $+1.1\%$/$+0.3\%$                                   \\
                                        & $(200, 200)$ & $-0.06$    &      $(11.8k, 15.2k)$         &        $+2.5\%$/$+1.3\%$                       \\ \midrule
    \multirow{5}{*}{$(30k, 30k)$} & Baseline & \doubleunderline{$42.42$}              &    $(29.3k, 29.5k)$           &        $28.59$/$31.42$                                   \\ 
                                        & $(50, 50)$   & $+0.13$    &      $(22.0k, 26.1k)$         &        $+0.4\%$/$+0.1\%$                                   \\
                                        & $(100, 100)$ & $+0.42$    &      $(15.1k, 19.9k)$         &        $+2.0\%$/$+1.3\%$                                   \\
                                        & $(150, 150)$ & \underline{$+0.72$}    &      $(12.2k, 15.7k)$         &        $+3.6\%$/$+2.9\%$                                   \\
                                        & $(200, 200)$ & \underline{$+0.72$}    &      $(10.4k, 13.2k)$         &        $+5.3\%$/$+4.6\%$                       \\ \bottomrule      
    \end{tabular}}
    \caption{A comparison between the several suboptimal baseline BPE models and their trimmed counterparts on the larger EN--FR dataset.\label{tab:suboptimalbaseline_french}}
\end{table}

\subsection{Extremely Large Base Vocabulary}\label{ssec:large}

In Section \ref{ssec:larger_dataset}, we observed that starting with an extremely large base vocabulary and trimming heavily tended to recover some BLEU performance.
This could have important implications for model hyperparameter choices, as it may suggest a strategy to reduce sequence lengths while retaining a reasonable parameter count and downstream performance.
To wit, one could pick a target parameter count and/or expected sequence length and then choose an extremely large base vocabulary before trimming it until the desired parameter count is met or an expected sequence length is passed.

For both the original setting and the larger corpus setting, we find that this is not a viable strategy, as one of the parameters always suffers (that is, either performance is worse, parameter count is too large, or the sequence length grows too long).
Specifically, we experiment with $20k$/$20k$ and $30k$/$30k$ in the original setting, as well as $40k$/$40k$ with the larger corpus, shown in \autoref{tab:extremely_large_bpe}.

In all cases, we observe a large increase in BLEU as the trimming threshold is increased, followed by a large decrease.
This effect is more pronounced in the smaller DE--EN vocabulary case, but that is possibly because of the same reasons discussed in Section~\ref{ssec:larger_dataset}---that the larger dataset means that the trained subword embeddings are more robust, which dampens factors that would affect BLEU. 

As for the strategy of picking an extremely large vocabulary before very aggressively trimming, we find that it does not confer an advantage on our metrics.
Take for example the EN-DE $(\mathbb{B}_s, \mathbb{B}_t)$ = $(30k, 30k)$, $(\mathbb{T}_s, \mathbb{T}_t) = (100, 100)$ case from \autoref{tab:extremely_large_bpe} and the $(\mathbb{B}_s, \mathbb{B}_t)$ = $(10k, 10k)$, $(\mathbb{T}_s, \mathbb{T}_t) = (150, 150)$ case from \autoref{tab:suboptimalbaseline}. These two have roughly the same effective vocabulary sizes $(3.0k, 2.5k)$ and $(3.1k, 2.4k)$, so in order to confer an advantage, the $30k$/$30k$ model would have to have either higher BLEU or shorter sequence lengths.
The $30k$/$30k$ model has slightly better BLEU ($33.88$ vs $33.79$), but longer sequences (27.81/27.32 vs  26.34/25.54).
We see that across all hyperparameters, the $20k$/$20k$ and $30k$/$30k$ models do not come close to outperforming the smaller base models and their trimmed variants, and often vastly underperform them.
Furthermore, like Section~\ref{ssec:stunted}, when comparing between models that have the same effective vocabulary counts, the trimmed smaller-base models have shorter sequences across the board compared to the trimmed larger-base models.
Thus, we find that this vocabulary selection strategy is not helpful.

\begin{table}[t]
    \centering
    \resizebox{\columnwidth}{!}{%
    \begin{tabular}{cc|ccc} \toprule
    \makecell{Vocabulary \\ $(\mathbb{B}_s, \mathbb{B}_t)$} & \makecell{Thresholds \\ $(\mathbb{T}_s, \mathbb{T}_t)$} & BLEU & \makecell{Effective \\ Vocabulary \\ $(\hat{\mathbb{B}}_{s}, \hat{\mathbb{B}}_{t})$} & \makecell{Sequence \\ Length \\ source/target} \\ \midrule
    \multirow{5}{*}{$(20k, 20k)$} & Baseline & \doubleunderline{$33.40$}             &    $(19.9k, 19.4k)$            &        $20.67$/$20.26$                                  \\ 
                                        & $(50, 50)$ &   \underline{$+0.50$}   &      $(6.4k, 4.8k)$            &        $+13.6\%$/$+13.3\%$                                   \\
                                        & $(100, 100)$ & $+0.49$   &      $(3.4k, 2.8k)$            &        $+27.3\%$/$+25.4\%$                                   \\
                                        & $(150, 150)$ & $+0.45$   &      $(2.4k, 1.9k)$            &        $+38.7\%$/$+37.7\%$                                   \\
                                        & $(200, 200)$ & $+0.04$   &      $(1.8k, 1.6k)$            &        $+49.4\%$/$+48.3\%$                       \\  \midrule
    \multirow{5}{*}{$(30k, 30k)$} & Baseline           & \doubleunderline{$32.79$}   &      $(29.6k, 28.1k)$          &        $20.16$/$19.99$                                   \\ 
                                        & $(50, 50)$   & $+0.82$   &      $(5.6k, 4.4k)$            &        $+20.6\%$/$+19.6\%$                                   \\
                                        & $(100, 100)$ & \underline{$+1.09$}   &      $(3.0k, 2.5k)$            &        $+37.9\%$/$+36.7\%$                                   \\
                                        & $(150, 150)$ & $+0.59$   &      $(2.1k, 1.8k)$            &        $+51.3\%$/$+54.9\%$                                   \\
                                        & $(200, 200)$ & $+0.26$   &      $(1.6k, 1.5k)$            &        $+63.2\%$/$+68.0\%$                       \\ \bottomrule
    \multicolumn{5}{c}{ (a) German\textrightarrow{}English } \\
    \toprule
    \makecell{Vocabulary \\ $(\mathbb{B}_s, \mathbb{B}_t)$} & \makecell{Thresholds \\ $(\mathbb{T}_s, \mathbb{T}_t)$} & BLEU & \makecell{Effective \\ Vocabulary \\ $(\hat{\mathbb{B}}_{s}, \hat{\mathbb{B}}_{t})$} & \makecell{Sequence \\ Length \\ source/target} \\ \midrule
    \multirow{11}{*}{$(30k, 30k)$} & Baseline & \doubleunderline{$42.42$}             &    $(29.3k, 29.5k)$          &        $28.59$/$31.42$                                  \\
                                        & $(50, 50)$ & $+0.13$      &      $(22.1k, 26.1k)$        &        $+0.4\%$/$+0.1\%$                                   \\
                                        & $(100, 100)$ & $+0.42$    &      $(15.1k, 20.0k)$        &        $+2.2\%$/$+1.3\%$                                   \\
                                        & $(150, 150)$ & $+0.72$    &      $(12.2k, 15.7k)$        &        $+3.6\%$/$+2.9\%$                                   \\
                                        & $(200, 200)$ & $+0.72$    &      $(10.4k, 13.2k)$        &        $+5.3\%$/$+4.6\%$                       \\
                                        & $(250, 250)$ & $+0.77$    &      $(9.3k, 11.5k)$         &        $+6.7\%$/$+6.2\%$                                 \\
                                        & $(300, 300)$ & $+0.78$    &     $(8.5k, 10.3k)$          &        $+8.2\%$/$+7.8\%$                                   \\
                                        & $(350, 350)$ & $+0.68$    &      $(7.9k, 9.4k)$          &        $+9.7\%$/$+9.1\%$                                   \\ 
                                        & $(400, 400)$ & $+0.85$    &      $(7.4k, 8.7k)$          &        $+10.9\%$/$+10.4\%$                                   \\ 
                                        & $(450, 450)$ & \underline{$+0.95$}    &      $(6.9k, 8.1k)$          &        $+12.4\%$/$+12.0\%$                                   \\ 
                                        & $(500, 500)$ & $+0.91$    &      $(6.5k, 7.6k)$          &        $+14.0\%$/$+13.3\%$                                   \\ \midrule
    \multirow{11}{*}{$(40k, 40k)$} & Baseline & \doubleunderline{$41.71$}             &      $(38.5k, 39.0k)$        &        $28.44$/$31.20$                                  \\ 
                                        & $(50, 50)$ & $+0.53$      &      $(20.0k, 26.6k)$        &        $+1.3\%$/$+0.8\%$                                 \\
                                        & $(100, 100)$ & $+1.14$    &      $(14.1k, 18.1k)$        &        $+3.3\%$/$+2.8\%$                                   \\
                                        & $(150, 150)$ & $+1.46$    &      $(11.5k, 14.4k)$        &        $+5.2\%$/$+4.8\%$                                   \\
                                        & $(200, 200)$ & $+1.56$    &      $(10.0k, 12.3k)$        &        $+7.3\%$/$+6.8\%$                       \\
                                        & $(250, 250)$ & $+1.69$    &      $(8.9k, 10.8k)$         &        $+9.0\%$/$+8.7\%$                       \\ 
                                        & $(300, 300)$ & \underline{$+1.75$}    &      $(8.2k, 9.7k)$          &        $+10.8\%$/$+10.4\%$                       \\
                                        & $(350, 350)$ & $+1.66$    &      $(7.6k, 8.9k)$          &        $+12.8\%$/$+12.2\%$                                   \\
                                        & $(400, 400)$ & $+1.53$    &      $(7.1k, 8.3k)$          &        $+15.4\%$/$+14.2\%$                                   \\
                                        & $(450, 450)$ & $+1.60$    &      $(6.7k, 7.8k)$          &        $+17.2\%$/$+16.3\%$                                   \\
                                        & $(500, 500)$ & $+1.54$    &      $(6.3k, 7.3k)$          &        $+18.9\%$/$+17.9\%$                                   \\ \bottomrule
                                        
        \multicolumn{5}{c}{ (b) English\textrightarrow{}French } \\
    \end{tabular}}
    \caption{Initializing an extremely large vocabulary and trimming for both (a) DE--EN and (b) EN--FR.\label{tab:extremely_large_bpe}}
\end{table}

\section{Discussion: Historical Benefits of Vocabulary Trimming}

Our experiments focus on modern neural machine translation, which has coalesced around the transformer architecture.
At the time of \newcite{sennrich-etal-2016-neural}, recurrent neural networks, convolutional networks, and even non-neural models were still popular choices for machine translation. Thus, the modeling best practices at the time may not apply to the more robust models used now.

Another aspect where trimming was useful is parameter reduction.
When subword modeling was proposed, it was not unusual to see models with vocabulary sizes of more than 80$k$~\cite{DBLP:conf/nips/SutskeverVL14, jean-etal-2015-montreal, sennrich-etal-2017-university}.
Coupled with the relatively small sizes of recurrent models, the embedding and decoding layers took up a disproportionately large amount of space, particularly with such large vocabularies~\cite{britz-etal-2017-massive}.
More recently, transformer models in NMT tend towards lower vocabulary sizes, and generally, smaller vocabularies means that the effect of vocabulary trimming is lessened.
Even in settings where larger vocabularies are used (for example, BERT- and GPT-style models, with vocabulary sizes in the 30-60$k$ range), the model's internal parameters dominate the overall parameter count.
These differences also manifest in the runtime efficiency of RNNs vs transformers.
Whereas in RNNs the simpler internal layers were faster to compute and runtime scaled linearly with sequence length so a large softmax layer is a performance bottleneck~\cite{devlin-etal-2014-fast, DBLP:journals/corr/GraveJCGJ16}, the runtime of transformers is dominated by the more complex, quadratic-runtime attention layers.

Modeling quality aside, for these practical reasons, vocabulary trimming, which decreases the model's parameter count and reduces the output softmax layer's runtime, was a useful optimization in the era of smaller, simpler recurrent architectures.
In the era of larger, deeper, more complex transformers, the relative benefits of this optimization are diminished due to other factors in the model.

\section{Related Work}

\paragraph{Subword Language Modeling Variants}
BPE is not the only subword language modeling technique.
Wordpiece~\cite{wordpiece} uses a similar vocabulary construction technique to BPE, but during inference it greedily selects the longest matching subword from the vocabulary.
Some techniques have sought to inject randomness as a data augmentation and regularizer.
BPE-Dropout~\cite{provilkov-etal-2020-bpe} uses the same initialization as BPE, but during inference it randomly prohibits merges, effectively causing the tokenizer to produce a distribution of tokenizations when given the same input.
This technique essentially eliminates the rare-subword issue but is also incompatible with subword trimming.
UnigramLM~\cite{kudo-2018-subword} is another stochastic tokenizer which builds a vocabulary by first creating a very large subword list and then iteratively pruning it according to some metric until the desired vocabulary size is reached.

Vocabulary construction is not the only tunable aspect of a tokenizer, as the \textit{inference} algorithm can be chosen as well.
For example, a vocabulary could be initialized with BPE but tokenization could be done via MaxMatch~\cite{wordpiece}, or by longest token~\cite{hofmann-etal-2022-embarrassingly}.
This not only influences sequence length and modeling quality, but can also affect the trimming procedure~\cite{uzan2024greed}.

Another choice is character-level modeling~\cite{clark-etal-2022-canine,xue-etal-2022-byt5}, which does not have an OOV or rare-subword problem, but can cause issues for transformers due to the extremely long sequences required.
For an overview of subword tokenization, see \newcite{DBLP:journals/corr/abs-2112-10508}.

\paragraph{Tokenizer Evaluation}
A number of works have considered the problem of tuning the vocabulary (and, thus, the embedding and decoding layers of a model) to ensure good performance. 
\newcite{galle-2019-investigating} evaluate subword tokenization techniques and observe that, holding vocabulary size constant, \textit{sequence compression} is a strong signal on the effectiveness of the subword tokenizer for a task.
This closely relates to our setting in Section \ref{ssec:stunted}, as the trimmed tokenizers usually produce longer sequences than untrimmed tokenizers with the same effective vocabulary size.
\newcite{gowda-may-2020-finding} recommend picking the largest vocabulary size where all tokens appear more than 100 times in the training corpus to ensure robustness, which is closely related to trimming.
We investigated trimming based on this heuristic in Section~\ref{ssec:100heuristic}.
\newcite{zouhar-etal-2023-tokenization} compare a number of intrinsic heuristics for evaluating tokenizers.
They find that an entropy-based metric, Rényi Efficiency, correlates best with downstream performance, followed closely by that of \newcite{gowda-may-2020-finding}.
Sequence compression was found to be much more weakly correlated with downstream performance.
Others investigate linguistic-inspired metrics such as morphological alignment~\cite{klein-tsarfaty-2020-getting, gow-smith-etal-2022-improving} or cognitive plausibility~\cite{beinborn-pinter-2023-analyzing}.

\section{Conclusion}

We present a thorough investigation, the first of its kind, of the commonplace practice to trim BPE tokenizer vocabularies in MT models based on the effective frequency of the tokens in the working corpus.
We show through extensive evaluation of ten specific hypotheses that for the well-researched corpus of IWSLT14 German\textrightarrow English this practice has little-to-no positive effect in modeling performance, despite its intuitive allure.
These hypotheses include restricting trimming to one side of the data or to nonterminal tokens, restricting evaluation to only rare subwords, and extending the scope of the initialization vocabulary.
The results hold for both the separate and joint vocabulary settings, as well as for a larger English\textrightarrow French task.

Our experimentation setup allows controlled analysis of the reasoning often associated with efficiency in subword-based models.
Overall, while we observe a slight reduction in model parameter count, as expected, it appears that the benefit this reduction affords transformer-based NMT models is limited and comes with an increase in sequence length.
We conclude that in the era of large, attention-based LLMs, there is no substantial advantage to subword trimming.

Even though the practice of vocabulary trimming is not ubiquitous in monolingual classification and generation settings, we believe the community will be well serviced by a complementary analysis for such settings.
Further inquiries into the prospects of vocabulary manipulation in other subword tokenization algorithms in various contexts might also prove useful.

\section*{Acknowledgements}
\textbf{Marco Cognetta and Naoaki Okazaki}: These research results were obtained from the commissioned research (No.22501) by National Institute of Information and Communications Technology (NICT) , Japan.

\noindent
\textbf{Tatsuya Hiraoka}: This work was supported by JST, ACT-X Grant Number JPMJAX21AM, Japan.

\noindent
\textbf{Yuval Pinter}: This research was supported in part by the Israel Science Foundation (grant No. 1166/23).

\bibliographystyle{acl_natbib}

\clearpage

\appendix

\section{Architecture and Training Details}
\label{apx:architecture}
\subsection{Architecture}
The same underlying \texttt{transformer-iwslt} architecture was used in all experiments.
\autoref{tbl:arch} gives the architecture and training details. In all experiments, only the embedding and decoding tables were changed between models from the same experiment batch, and all other aspects of the underlying language model architecture were held constant. 

\begin{table}[H]
    \centering
    \begin{tabular}{lc}\toprule
        FFN Dim &  $1024$            \\
        Embedding Dim &  $512$       \\
        \#Heads &  $4$               \\ 
        Encoder Layers  &  $6$       \\ 
        Decoder Layers  &  $6$       \\ \midrule
        Tokens Per Batch &  $4096$   \\
        Optimizer &  ADAM            \\
        Learning Rate &  5$e$-4      \\
        Betas &  $(0.9, 0.98)$       \\
        Learning Rate Scheduler &  \texttt{inverse sqrt}      \\
        Warm-up Steps &  4000      \\
        \bottomrule
    \end{tabular}
    \caption{The architecture and training details used in all EN-DE experiments. The vocabulary size is omitted, as we vary that in each experiment.}
    \label{tbl:arch}
\end{table}

\begin{table}[H]
    \centering
    \begin{tabular}{lc}\toprule
        FFN Dim &  $2048$            \\
        Embedding Dim &  $512$       \\
        \#Heads &  $8$               \\ 
        Encoder Layers  &  $6$       \\ 
        Decoder Layers  &  $6$       \\ \midrule
        Tokens Per Batch &  $4096$   \\
        Optimizer &  ADAM            \\
        Learning Rate &  5$e$-4      \\
        Betas &  $(0.9, 0.98)$       \\
        Learning Rate Scheduler &  \texttt{inverse sqrt}      \\
        Warm-up Steps &  4000      \\
        \bottomrule
    \end{tabular}
    \caption{The architecture and training details used in all FR-EN experiments. The vocabulary size is omitted, as we vary that in each experiment.}
    \label{tab:large_arch}
\end{table}

\newpage
\newpage
\section{Rare-Word Sentences}\label{app:rare}
\begin{table*}[h]
    \centering
    \resizebox{1.9\columnwidth}{!}{%
    \begin{tabular}{cc|cccccc} \toprule
    \makecell{Vocabulary \\ $(\mathbb{B}_s, \mathbb{B}_t)$} & \makecell{Thresholds \\ $(\mathbb{T}_s, \mathbb{T}_t)$} & \makecell{BLEU\\(overall)} & \makecell{BLEU\\(source mismatch) \\ Baseline/Trimmed} & \makecell{BLEU\\(source match) \\ Baseline/Trimmed} & \makecell{BLEU\\(target mismatch) \\ Baseline/Trimmed} & \makecell{BLEU\\(target match) \\ Baseline/Trimmed} & \makecell{BLEU\\(both mismatch) \\ Baseline/Trimmed} \\ \midrule
    \multirow{16}{*}{($6k$, $6k$)} & Baseline     & $34.05$    &      -           &        -                             & - & - & -    \\ \
                                     & (0, 100) & $33.88$ & -   &      -           &        $32.22$/$32.05$                             & $36.50$/$36.32$ & -     \\
                                     & (0, 150) & $33.98$ & -   &      -           &        $33.04$/$33.11$                             & $38.21$/$37.58$ & -     \\
                                     & (0, 200) & $33.92$ & -   &      -           &        $33.11$/$33.07$                             & $40.07$/$39.53$ & -     \\
                                     & (100, 0) & $34.10$ & $31.92$/$31.91$   &      $34.95$/$35.04$           &        -                             & - & -     \\
                                     & (150, 0) & $33.82$ & $33.02$/$32.76$   &      $37.33$/$37.17$           &        -                             & - & -     \\
                                     & (200, 0) & $33.92$ & $33.38$/$33.21$   &      $38.91$/$38.98$           &        -                             & - & -     \\
                                     & (100, 100) & $33.77$ & $31.92$/$31.65$   &      $34.95$/$34.68$           &        $32.22$/$32.00$                             & $36.50$/$36.16$  & $30.96$/$30.70$    \\
                                     & (100, 150) & $33.39$ & $31.92$/$31.11$   &      $34.95$/$34.35$           &        $33.04$/$32.42$                             & $38.21$/$37.36$  & $31.67$/$30.80$    \\
                                     & (100, 200) & $33.64$ & $31.92$/$31.48$   &      $34.95$/$34.56$           &        $33.11$/$32.78$                             & $40.07$/$39.16$  & $31.82$/$31.39$    \\
                                     & (150, 100) & $33.78$ & $33.02$/$32.06$   &      $37.33$/$36.15$           &        $32.22$/$31.25$                             & $36.50$/$35.43$  & $31.92$/$31.87$    \\
                                     & (150, 150) & $33.77$ & $31.58$/$32.70$   &      $37.33$/$37.13$           &        $31.87$/$32.85$                             & $37.40$/$37.58$  & -    \\
                                     & (150, 200) & $33.83$ & $33.02$/$32.86$   &      $37.33$/$36.94$           &        $33.11$/$32.94$                             & $40.07$/$39.55$  & $32.77$/$32.65$    \\
                                     & (200, 100) & $33.83$ & $33.38$/$33.11$   &      $38.91$/$38.96$           &        $32.22$/$32.07$                             & $36.50$/$36.17$  & $31.93$/$31.75$    \\
                                     & (200, 150) & $33.93$ & $33.38$/$33.28$   &      $38.91$/$38.58$           &        $33.04$/$33.00$                             & $38.21$/$37.74$  & $32.78$/$32.88$    \\
                                     & (200, 200) & $33.75$ & $33.38$/$33.15$   &      $38.91$/$38.74$           &        $33.11$/$32.90$                             & $40.07$/$39.76$  & $33.01$/$32.95$    \\ \midrule
    \multirow{16}{*}{($10k$, $10k$)} & Baseline & $33.56$    &      -           &        -                             & - & - & -    \\ \
                                     & (0, 100) & $33.44$ & -   &      -           &        $32.35$/$32.24$                             & $38.55$/$38.36$  & -    \\
                                     & (0, 150) & $33.83$ & -   &      -           &        $32.87$/$32.99$                             & $39.67$/$39.93$  & -    \\
                                     & (0, 200) & $33.66$ & -   &      -           &        $32.81$/$32.94$                             & $41.08$/$40.91$  & -    \\
                                     & (100, 0) & $33.58$ & $32.93$/$32.93$   &      $36.74$/$36.71$           &        -                             & -  & -    \\
                                     & (150, 0) & $33.47$ & $33.10$/$33.02$   &      $38.10$/$37.82$           &        -                             & -  & -    \\
                                     & (200, 0) & $33.50$ & $33.22$/$33.11$   &      $38.99$/$39.42$           &        -                             & -  & -    \\
                                     & (100, 100) & $33.93$ & $32.93$/$33.30$   &      $36.74$/$37.03$           &        $32.35$/$32.71$                             & $38.55$/$38.98$   & $32.44$/$32.81$   \\
                                     & (100, 150) & $33.86$ & $32.93$/$33.26$   &      $36.74$/$36.90$           &        $32.73$/$33.03$                             & $39.75$/$40.00$   & $32.87$/$33.14$   \\
                                     & (100, 200) & $33.70$ & $31.08$/$33.06$   &      $36.74$/$36.87$           &        $31.18$/$32.98$                             & $40.38$/$41.08$   & $32.48$/$33.19$   \\
                                     & (150, 100) & $33.70$ & $33.10$/$33.22$   &      $38.10$/$38.17$           &        $32.35$/$32.52$                             & $38.55$/$38.61$   & $32.34$/$32.50$   \\
                                     & (150, 150) & $33.79$ & $33.10$/$33.29$   &      $38.10$/$38.57$           &        $32.73$/$32.98$                             & $39.75$/$39.74$   & $32.73$/$32.98$   \\
                                     & (150, 200) & $33.80$ & $33.10$/$33.31$   &      $38.10$/$38.40$           &        $32.81$/$33.05$                             & $41.08$/$41.14$   & $32.79$/$33.04$   \\
                                     & (200, 100) & $33.87$ & $33.22$/$33.49$   &      $38.99$/$39.65$           &        $32.35$/$32.66$                             & $38.55$/$38.79$   & $32.40$/$32.74$   \\
                                     & (200, 150) & $33.74$ & $33.22$/$33.38$   &      $38.99$/$39.22$           &        $32.73$/$32.92$                             & $39.75$/$39.64$   & $32.72$/$32.88$   \\
                                     & (200, 200) & $33.74$ & $33.22$/$33.35$   &      $38.99$/$39.56$           &        $32.81$/$32.97$                             & $41.08$/$41.08$   & $32.81$/$32.99$   \\ \midrule
    \multirow{16}{*}{($12k$, $12k$)} & Baseline   & $33.68$    &      -           &        -                             & - & -  & -   \\ \
                                       & (0, 100) & $33.71$ & -   &      -           &        $32.54$/$32.54$                             & $39.36$/$39.43$  & -    \\
                                       & (0, 150) & $33.75$ & -   &      -           &        $32.84$/$32.92$                             & $40.37$/$40.34$  & -    \\
                                       & (0, 200) & $33.77$ & -   &      -           &        $32.93$/$33.08$                             & $41.03$/$41.29$  & -    \\
                                       & (100, 0) & $33.69$ & $33.08$/$33.09$   &      $38.03$/$38.01$           &        -                             & -  & -    \\
                                       & (150, 0) & $33.59$ & $33.32$/$33.22$   &      $38.50$/$38.58$           &        -                             & -  & -    \\
                                       & (200, 0) & $33.51$ & $33.29$/$33.21$   &      $39.80$/$40.28$           &        -                             & -  & -    \\
                                       & (100, 100) & $33.83$ & $32.87$/$33.37$   &      $37.90$/$38.00$           &        $32.40$/$32.81$                             & $38.80$/$39.39$  & $32.54$/$32.73$    \\
                                       & (100, 150) & $33.77$ & $33.23$/$33.28$   &      $37.96$/$38.37$           &        $32.91$/$33.03$                             & $40.89$/$40.67$  & $32.75$/$32.80$    \\
                                       & (100, 200) & $33.61$ & $33.08$/$33.00$   &      $38.03$/$38.11$           &        $32.96$/$32.89$                             & $41.35$/$41.15$  & $32.85$/$32.80$    \\
                                       & (150, 100) & $34.08$ & $33.32$/$33.68$   &      $38.50$/$39.41$           &        $32.54$/$33.00$                             & $39.36$/$39.40$  & $32.57$/$33.00$    \\
                                       & (150, 150) & $33.61$ & $33.25$/$33.10$   &      $38.56$/$38.78$           &        $32.81$/$32.68$                             & $40.11$/$40.12$  & $32.85$/$32.71$    \\
                                       & (150, 200) & $33.67$ & $33.47$/$33.29$   &      $38.39$/$39.19$           &        $33.03$/$32.95$                             & $42.00$/$41.52$  & $32.97$/$32.92$    \\
                                       & (200, 100) & $33.96$ & $33.47$/$33.62$   &      $39.56$/$40.21$           &        $32.60$/$32.83$                             & $39.64$/$39.54$  & $32.50$/$32.76$    \\
                                       & (200, 150) & $33.79$ & $33.36$/$33.44$   &      $39.69$/$39.91$           &        $32.84$/$32.92$                             & $40.37$/$40.62$  & $32.86$/$32.94$    \\
                                       & (200, 200) & $33.84$ & $33.29$/$33.55$   &      $39.80$/$39.63$           &        $32.93$/$33.14$                             & $41.03$/$41.35$  & $32.97$/$33.09$    \\ \bottomrule
    \end{tabular}
    }
    \caption{A comparison of baseline models and trimmed models on a test set composed of sentences that contain rare subwords (subwords that would be trimmed from the baseline model if threshold $\mathbb{T}$ was used). That is, suppose we have a baseline tokenizer $\mathcal{A}$ and a trimmed tokenizer $\mathcal{A}'$. Given a sentence pair $(s,t)$, if the tokenization of $s$ differs between $\mathcal{A}$ and $\mathcal{A}'$, we add it to a \say{mismatched source} test set. The same is repeated for each of $\{$source, target$\} \times \{$match, mismatch$\}$. In each case, we report the BLEU score for both the baseline model and the trimmed model (shown as Baseline/Trimmed) that were trained with those tokenizers. For configurations where $\mathbb{T}_s = 0$, the ``source mismatched'' test set is empty (since the source tokenizer is unchanged) and the ``source matched'' test set is equal to the overall test set, and analogously for $\mathbb{T}_t = 0$ and target sets. Thus, we omit these values.\label{tab:raresentences}}
\end{table*}

\end{document}